\documentclass{article}
\usepackage[T1]{fontenc}
\usepackage{spconf,amsmath,booktabs,graphicx,hyperref,placeins,float}
\hypersetup{hidelinks}
\let\originalthebibliography\thebibliography
\renewcommand{\thebibliography}[1]{%
  \originalthebibliography{#1}%
  \setlength{\itemsep}{0pt}%
  \setlength{\parsep}{0pt}%
}

\title{HearInContext: A Benchmark for Implicit Context in Speech Recognition}
\name{Yifan Gao, Yao Tian, Hongbin Suo\sthanks{Corresponding author.}}
\address{
  AI Center, OPPO, Beijing, China\\
  \texttt{\{gaoyifan,aaron1,suohongbin\}@oppo.com}
}

\begin{document}
\maketitle
\begin{abstract}
Contextual ASR can benefit from semantic cues or from target words
explicitly provided in the context. We introduce HearInContext,
a Mandarin--English benchmark that pairs shared synthetic speech
with assistant replies supporting different interpretations.
The benchmark comprises 3,764 semantic test cases built around homophones.
Implicit contexts exclude candidate words; explicit contexts name
the target. No-context and unrelated-context controls measure
the benefit of relevant history and sensitivity to irrelevant history.
Context-capable models benefit from implicit cues but achieve higher
target recall with explicit hints. Fine-tuning Qwen3-ASR-1.7B improves
implicit-context target recall by 11.4 percentage points
in both Mandarin and English, while absolute CER/WER
changes on AISHELL-1 and LibriSpeech remain below 0.1 percentage
points. Gains extend to explicit conditions excluded from fine-tuning
and to Mandarin hotword recognition on real recordings.
Code and data are available at
\url{https://github.com/OPPO-Mente-Lab/HearInContext}.
\end{abstract}

\section{Introduction}
\label{sec:intro}
Conversational context can resolve ambiguities that speech alone
cannot. For example, a discussion of baking favors \emph{flour},
whereas one about choosing a rose as a gift favors \emph{flower}, even when neither
word has been mentioned. Resolving such ambiguity requires an ASR
model to connect the meaning of the preceding conversation with
the current utterance.

\begin{figure*}[t]
\centering
\includegraphics[width=0.94\textwidth]{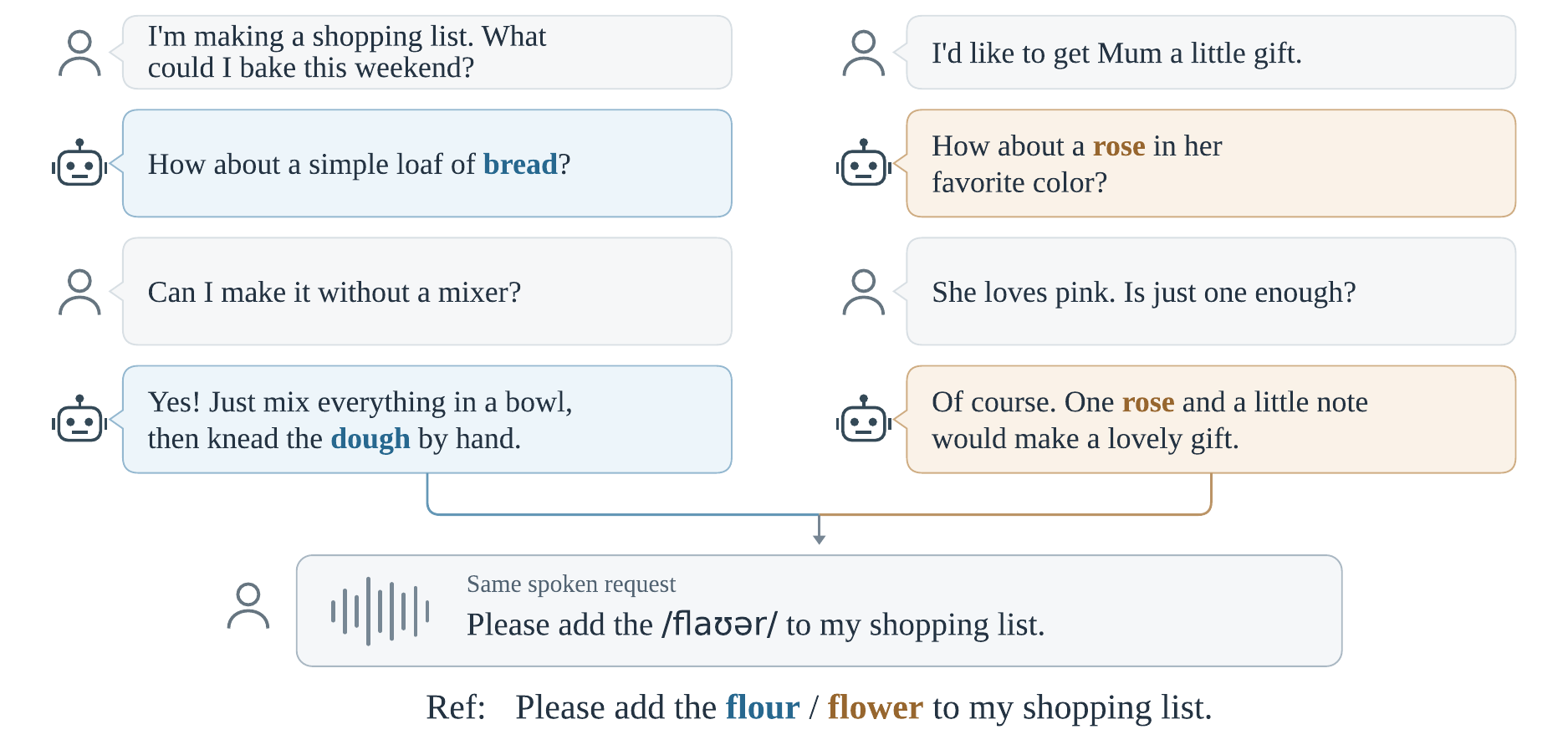}
\caption{Illustrative example of implicit contextual disambiguation.
The same audio supports \emph{flour} or \emph{flower} under different
assistant histories, neither of which names the candidate words.
The dialogue is illustrative rather than a test-set excerpt;
the waveform is schematic.}
\label{fig:design}
\end{figure*}
Existing approaches draw on bias phrases \cite{deepcontext,salm,speechllm_biasing},
domain prompts \cite{domainprompt}, and conversational history
\cite{hori2021context,mu2026mars,zheng2026multilevel}.
DANCER uses entity descriptions to resolve phonetic confusion
in transcript correction \cite{dancer}, while prompt diagnostics
examine whether models follow the intended instructions \cite{doprompts}.
Mohebbi et al.\ \cite{mohebbi2023} probe homophone resolution
using utterance-internal syntactic cues; we examine preceding
conversational semantics while controlling candidate-word exposure.
We distinguish between \emph{explicit} hints that name
the target and \emph{implicit} context that supports its meaning
without naming any candidate.

Existing benchmarks provide documents and slides \cite{conec,slidespeech},
domain and entity hints \cite{contextasrbench}, professional profiles
\cite{profasr}, or descriptions and entity lists \cite{indiccontexteval}.
Recognition gains alone do not reveal whether a model uses semantic
context or benefits from seeing the target word.
Without shared audio, differences in recording quality can also affect
the comparison. Holding the waveform fixed lets us test whether
recognition changes with the supported interpretation.

We introduce \textbf{HearInContext}, a benchmark for implicit contextual
disambiguation with a controlled test set and separate training and
development sets. It distinguishes these sources of
improvement by pairing the same audio with implicit, explicit, and
unrelated contexts, alongside a no-context baseline. Implicit histories
support competing interpretations without naming any candidate.
We use assistant replies: unlike prior spoken user turns, this
system-generated text is available without ASR errors.
An assistant can establish a topic or suggest a next action without
supplying the user's eventual wording.

Adaptation experiments with Qwen3-ASR-1.7B \cite{qwen3asr} examine
whether implicit recognition gains can coexist with general ASR
performance and robustness to unrelated context.
Our main contributions are:
\begin{itemize}
\setlength{\itemsep}{1pt}
\setlength{\parsep}{0pt}
\item We introduce a bilingual benchmark that separates implicit
semantic support from explicit word hints through candidate-word
exclusion and shared-audio comparisons.
\item We characterize the gap between implicit and explicit context
use through model comparisons and context-source ablations.
\item We demonstrate implicit disambiguation gains and transfer to
explicit hints and Mandarin hotwords on real speech, while testing
general recognition and robustness to unrelated context.
\end{itemize}

\newpage
\section{HearInContext}
\label{sec:benchmark}
\begin{table}[H]
\centering
\caption{Corpus statistics before resampling. Audio counts and hours refer to unique waveforms.}
\label{tab:dataset}
\small
\renewcommand{\arraystretch}{1.08}
\setlength{\tabcolsep}{2pt}
\begin{tabular*}{\columnwidth}{@{\extracolsep{\fill}}llrrrr@{}}
\toprule
Split & Lang. & Cases & Audio & Hours & \shortstack{Implicit\\length} \\
\midrule
Train & ZH & 5,400 & 10,800 & 17.25 & 389.4 \\
      & EN & 1,800 & 3,600  & 4.20  & 354.9 \\
\addlinespace[2pt]
Dev   & ZH & 600 & 1,200 & 1.92 & 389.2 \\
      & EN & 200 & 400   & 0.48 & 360.8 \\
\addlinespace[2pt]
Test  & ZH & 2,656 & 2,618 & 2.35 & 316.8 \\
      & EN & 1,108 & 1,100 & 0.81 & 210.3 \\
\bottomrule
\end{tabular*}
\par\smallskip
\begin{minipage}{\columnwidth}
\small
Implicit length: mean non-whitespace characters (ZH) or words (EN).
Each test case is evaluated under No context, Implicit, Explicit, and
Unrelated conditions, yielding 7,528 case--voice examples per condition
with shared audio.
\end{minipage}
\end{table}

\subsection{Benchmark Design}
A group shares audio across competing homophones in one sentence
frame. Each case pairs a candidate with its history and reference
(Figure~\ref{fig:design}).

Assistant replies form the default context; user turns are retained
for source ablations.
Each case uses the same waveform and reference across four conditions:
\emph{No context} supplies no history;
\emph{Implicit} supports the target without naming any candidate;
\emph{Explicit} names the target; and \emph{Unrelated} supplies
irrelevant history. These conditions test whether context helps,
explicit hints add further benefit, and irrelevant history interferes.
One audio-only transcription cannot match every branch reference:
no-context scores measure unresolved ambiguity, while separate corpora
assess general recognition.

\subsection{Data Construction}
\textbf{Candidate groups and target utterances.}
Each candidate group contains homophones with distinct meanings
that fit a common sentence frame. We synthesize one anchor utterance
per group and reuse its audio across semantic branches.

\textbf{Dialogue construction and review.}
We construct assistant histories supporting each interpretation,
then derive explicit and unrelated controls. Explicit contexts name
the target in one reply; unrelated contexts use lexically dissimilar
histories from another group and domain in the same language.
Automated checks cover both languages: references contain the target
once; implicit and unrelated contexts exclude all candidates;
explicit contexts contain only the intended candidate.
English groups additionally undergo DeepSeek-V4-flash\footnote{\url{https://api-docs.deepseek.com/}} review for pronunciation--sense agreement,
sentence naturalness, and contextual support.
Explicit rewrites retain other replies and temporal coherence
without reproducing the full reference sentence.

\textbf{Speech synthesis.}
CosyVoice2-0.5B \cite{cosyvoice2} uses disjoint pools of 96 adaptation
and 26 test reference speakers, each balanced equally by gender.
Mandarin references come from AISHELL-3 \cite{aishell3} and English
references from VCTK \cite{vctk}. Each test group uses one male and
one female reference speaker from its language-specific pool to
synthesize two versions of the anchor utterance.
The 16-kHz waveforms are reused across semantic branches and
conditions after format and metadata checks.

\subsection{Dataset Composition and Splits}
The 1,859 test groups contain 2,656 Mandarin and 1,108 English
cases. Two recordings per group yield 3,718 waveforms and 7,528 case--voice
examples per condition (Table~\ref{tab:dataset}).

A separate adaptation corpus contains 7,200 training and 800
development targets with homophonic or near-homophonic competitors.
Each target sentence is synthesized separately in two voices,
unlike the shared-audio test branches. Implicit contexts are generated
and reviewed by DeepSeek-V4-flash and checked for target-word exclusion.
Training/development targets are lexically disjoint from test targets.

\section{Experimental Setup}
\label{sec:setup}
\subsection{Models and Decoding}
We compare FireRedASR2-AED \cite{firered2}, SenseVoice-Small
\cite{sensevoice}, and Whisper-Large-v3 \cite{whisper} without
context. Context-aware evaluation includes Qwen3-ASR-0.6B and
1.7B \cite{qwen3asr}, VibeVoice-ASR-7B \cite{vibevoiceasr}, and
Seed-ASR \cite{seedasr}, accessed through the Seed-ASR 2.0 API.
We use native context interfaces; on HearInContext, Qwen
receives the history without an additional instruction.

Qwen uses vLLM \cite{vllm}, greedy generation, a 512-token output
limit, and the known language before and after fine-tuning.
The reproduction materials document remaining settings and output parsing.

\subsection{Supervised Adaptation}
We fine-tune all parameters of Qwen3-ASR-1.7B on the CosyVoice
training data. As implicit disambiguation is the primary adaptation
objective, we fix implicit examples at 80\% in the main experiment.
The remaining 20\% is split between empty and unrelated contexts
to compare recognition retention with robustness to irrelevant context.
We compare 80/20/0, 80/10/10, and 80/0/20 mixtures, ordered as
\emph{Implicit / No context / Unrelated}. Chinese and English
are sampled equally within each condition. Every configuration
contains 14,400 training examples; explicit contexts are evaluation-only.

All mixtures use AdamW, a learning rate of $10^{-5}$, linear decay
with 2\% warmup, and an effective batch size of 16.
Training runs for 900 optimizer updates, saving a checkpoint every
100 updates. Under the frozen development protocol, we select each
mixture's checkpoint by summed ascending error-rate and descending
recall ranks over six language--condition subsets, plus 0.25 times
summed ascending ranks of the corresponding across-speaker standard
deviations. Ties favor lower performance-rank sums, then earlier
steps, selecting 700, 800, and 400.

\subsection{Evaluation Protocol}
We report Mandarin CER, English WER, and target recall.
HearInContext counts at most one hit per example when the normalized
target occurs contiguously in the hypothesis.
References, hypotheses, and targets are independently normalized:
numeric normalization and character/Latin-word tokenization for
Mandarin; initialism processing and Whisper's EnglishTextNormalizer
for English. Scores pool edit counts or target hits across examples.

General ASR is evaluated without context on the official
AISHELL-1 test set \cite{aishell1} and LibriSpeech test-clean and
test-other \cite{librispeech}. ContextASR-Speech \cite{contextasrbench}
tests entity recall under coarse domain context. Real-speech hotword
transfer uses 808 Test-AISHELL1-NE utterances \cite{seaco}, each given
the same official 400-word list. All configurations use
greedy decoding with a 1,024-token limit and automatic language detection.
Keyword recall pools normalized contiguous hits capped by each keyword's
reference count. External sets are evaluation-only. Base source ablations
compare assistant, ground-truth user, and full histories on CosyVoice audio.

\section{Results and Analysis}
\label{sec:results}
\subsection{Implicit versus Explicit Context}
\vfill
\begin{table}[H]
\centering
\caption{Model comparison (\%). $R$: target recall.}
\label{tab:benchmark}
\small
\renewcommand{\arraystretch}{1.05}
\setlength{\tabcolsep}{2pt}
\begin{tabular*}{\columnwidth}{@{\extracolsep{\fill}}l*{3}{rr}@{}}
\toprule
\textbf{Chinese} & \multicolumn{2}{c}{No context} & \multicolumn{2}{c}{Implicit} & \multicolumn{2}{c}{Explicit} \\
\cmidrule(lr){2-3}\cmidrule(lr){4-5}\cmidrule(lr){6-7}
Model & CER$\downarrow$ & $R\uparrow$ & CER$\downarrow$ & $R\uparrow$ & CER$\downarrow$ & $R\uparrow$ \\
\midrule
FireRed2-AED & 8.23 & 44.77 & \multicolumn{4}{c}{--} \\
SenseVoice & 8.14 & 42.15 & \multicolumn{4}{c}{--} \\
Whisper-v3 & 10.54 & 40.59 & \multicolumn{4}{c}{--} \\
\addlinespace[2pt]
Qwen-0.6B & 7.88 & 44.95 & 5.29 & 65.66 & 3.54 & 79.29 \\
Qwen-1.7B & \textbf{7.73} & \textbf{45.41} & 4.52 & \textbf{71.80} & 3.52 & 83.23 \\
VibeVoice & 8.46 & 43.75 & \textbf{4.32} & 70.93 & \textbf{1.74} & \textbf{88.99} \\
SeedASR & 8.29 & 43.67 & 4.74 & 66.96 & 2.37 & 84.04 \\
\midrule
\textbf{English} & \multicolumn{2}{c}{No context} & \multicolumn{2}{c}{Implicit} & \multicolumn{2}{c}{Explicit} \\
\cmidrule(lr){2-3}\cmidrule(lr){4-5}\cmidrule(lr){6-7}
Model & WER$\downarrow$ & $R\uparrow$ & WER$\downarrow$ & $R\uparrow$ & WER$\downarrow$ & $R\uparrow$ \\
\midrule
FireRed2-AED & 11.77 & 43.59 & \multicolumn{4}{c}{--} \\
SenseVoice & 12.30 & 37.64 & \multicolumn{4}{c}{--} \\
Whisper-v3 & 10.44 & 43.86 & \multicolumn{4}{c}{--} \\
\addlinespace[2pt]
Qwen-0.6B & 10.21 & 43.28 & 6.46 & 66.20 & 3.02 & 87.45 \\
Qwen-1.7B & \textbf{10.09} & \textbf{44.00} & \textbf{5.15} & \textbf{73.69} & 2.26 & 90.75 \\
VibeVoice & 11.38 & 39.89 & 6.07 & 69.68 & \textbf{2.13} & \textbf{91.56} \\
SeedASR & 11.71 & 38.36 & 7.45 & 60.97 & 5.26 & 74.05 \\
\bottomrule
\end{tabular*}
\end{table}
\newpage

In Table~\ref{tab:benchmark}, dashes mark unevaluated contextual conditions
for the three no-context baselines. All four context-capable models
benefit from implicit cues in both languages. For Qwen3-ASR-1.7B,
target recall rises from 45.41\% to 71.80\% in Mandarin and from
44.00\% to 73.69\% in English. Yet explicit hints yield another
11.43 and 17.06 percentage points, respectively. Supplying a word
and inferring it from context therefore remain distinct challenges.

VibeVoice leads in explicit recall, whereas Qwen3-ASR-1.7B leads
in implicit recall in both languages. VibeVoice nevertheless has
lower Mandarin implicit CER, illustrating that overall transcription
quality and target recovery provide complementary information.

\subsection{Balancing Gains and Robustness}
\begingroup
\setlength{\intextsep}{6pt}
\begin{table}[H]
\centering
\caption{Qwen3-ASR-1.7B fine-tuning (\%). Mixture order: Implicit/No context/Unrelated.}
\label{tab:finetuning}
\small
\renewcommand{\arraystretch}{1.0}
\setlength{\tabcolsep}{2.5pt}
\begin{tabular*}{\columnwidth}{@{\extracolsep{\fill}}l*{4}{r}@{}}
\toprule
Metric & Base & 80/20/0 & 80/10/10 & 80/0/20 \\
\midrule
\multicolumn{5}{@{}l}{\textbf{HearInContext -- Chinese}} \\
No context CER $\downarrow$ & 7.73 & 7.67 & \textbf{7.65} & 7.73 \\
No context $R\uparrow$ & 45.41 & 45.56 & \textbf{45.58} & 45.39 \\
Implicit CER $\downarrow$ & 4.52 & \textbf{2.86} & 3.37 & 3.81 \\
Implicit $R\uparrow$ & 71.80 & \textbf{87.03} & 83.15 & 80.29 \\
Unrelated CER $\downarrow$ & 8.78 & 9.35 & \textbf{8.48} & \textbf{8.48} \\
Unrelated $R\uparrow$ & 43.17 & 38.08 & 43.02 & \textbf{44.20} \\
\midrule
\multicolumn{5}{@{}l}{\textbf{HearInContext -- English}} \\
No context WER $\downarrow$ & 10.09 & 9.63 & \textbf{9.56} & 9.64 \\
No context $R\uparrow$ & 44.00 & \textbf{44.31} & 44.22 & 43.64 \\
Implicit WER $\downarrow$ & 5.15 & \textbf{2.34} & 2.83 & 2.81 \\
Implicit $R\uparrow$ & 73.69 & \textbf{88.13} & 85.06 & 85.38 \\
Unrelated WER $\downarrow$ & 10.63 & 11.27 & 10.26 & \textbf{10.07} \\
Unrelated $R\uparrow$ & 41.38 & 37.05 & 41.38 & \textbf{42.64} \\
\midrule
\multicolumn{5}{@{}l}{\textbf{General ASR}} \\
AISHELL-1 CER $\downarrow$ & 1.51 & 1.52 & \textbf{1.50} & 1.52 \\
Libri clean WER $\downarrow$ & \textbf{1.63} & 1.70 & 1.68 & 1.69 \\
Libri other WER $\downarrow$ & 3.40 & 3.44 & \textbf{3.33} & 3.41 \\
\midrule
\multicolumn{5}{@{}l}{\textbf{ContextASR -- entity recall}} \\
ZH Coarse $\uparrow$ & 89.00 & \textbf{89.62} & 89.60 & 89.30 \\
EN Coarse $\uparrow$ & 85.77 & 84.66 & \textbf{86.46} & 83.85 \\
\midrule
\multicolumn{5}{@{}l}{\textbf{AISHELL-1-NE -- real speech}} \\
No hotwords CER $\downarrow$ & 4.25 & \textbf{4.21} & \textbf{4.21} & 4.26 \\
No hotwords $R\uparrow$ & 67.48 & \textbf{68.96} & 67.37 & 68.01 \\
Hotwords CER $\downarrow$ & 3.80 & 3.52 & \textbf{3.51} & 3.69 \\
Hotwords $R\uparrow$ & 73.09 & \textbf{75.85} & 74.15 & 74.68 \\
\bottomrule
\end{tabular*}
\end{table}
\endgroup

All mixtures improve implicit recognition (Table~\ref{tab:finetuning}).
The 80/20/0 mixture maximizes implicit recall but loses 5.08/4.33
points under unrelated context (ZH/EN), despite improved audio-only
error rates. Here, empty-context examples do not substitute for
unrelated-context examples during adaptation.

With 80/10/10, unrelated recall drops by 0.15 points in Mandarin
and stays unchanged in English; CER/WER improves. The 80/0/20 mixture gives higher
unrelated recall but lower Mandarin implicit recall; no mixture
dominates all conditions.

Does adaptation improve context use, or simply make target words
easier to recognize? For 80/10/10, no-context target recall changes from 45.41\% to
45.58\% in Mandarin and from 44.00\% to 44.22\% in English, whereas
implicit-context recall increases to 83.15\% and 85.06\%, respectively.
Consequently, relative to Base, the Implicit--No-context recall gap
widens by 11.18 and 11.15 percentage points. This pattern is consistent
with improved use of relevant context rather than a general increase
in target-word recovery.

Gains extend beyond the trained implicit condition: 80/10/10 raises explicit
recall from 83.23\% to 91.27\% in Mandarin and from 90.75\%
to 96.71\% in English, demonstrating cross-condition transfer.

For 80/10/10, AISHELL-1 and LibriSpeech error rates remain within
0.1 percentage points of Base. ContextASR-Speech coarse recall
improves by 0.60/0.69 points (ZH/EN).

All mixtures also improve hotword-conditioned recognition on
Test-AISHELL1-NE real recordings: 80/10/10 gives the lowest CER,
and 80/20/0 the highest recall (Table~\ref{tab:finetuning}).
\subsection{Evidence in Assistant Replies}
\begingroup
\setlength{\intextsep}{6pt}
\begin{table}[H]
\centering
\caption{Context sources with Qwen3-ASR-1.7B Base (\%).}
\label{tab:context-source}
\small
\renewcommand{\arraystretch}{1.08}
\setlength{\tabcolsep}{3pt}
\begin{tabular*}{\columnwidth}{@{\extracolsep{\fill}}lrrrr@{}}
\toprule
& \multicolumn{2}{c}{\textbf{Chinese}}
& \multicolumn{2}{c}{\textbf{English}} \\
\cmidrule(lr){2-3}\cmidrule(lr){4-5}
Context source & CER$\downarrow$ & $R\uparrow$ & WER$\downarrow$ & $R\uparrow$ \\
\midrule
Assistant-only & 4.52 & 71.80 & \textbf{5.15} & \textbf{73.69} \\
User-only & \textbf{4.25} & 70.12 & 5.70 & 70.85 \\
Full-history & 4.56 & \textbf{72.21} & 5.55 & \textbf{73.69} \\
\bottomrule
\end{tabular*}
\par\smallskip
\begin{minipage}{\columnwidth}
\small
User histories are ground truth; assistant replies are system text.
\end{minipage}
\end{table}
\endgroup

Table~\ref{tab:context-source} tests how much benefit remains without
prior user turns. Assistant-only context lowers error rates relative to
full history in both languages, with equal English recall and a
0.41-point loss in Mandarin recall. User-only has the lowest Mandarin
CER but uses ground-truth transcripts. Assistant replies thus offer
a practical default, retaining most full-history benefit without
transcribing prior user speech.

\FloatBarrier
\section{Conclusion}
HearInContext evaluates implicit contextual disambiguation by holding
audio fixed and excluding candidate words from the context.
Fine-tuning improves recognition with implicit and explicit context
and on a real-speech Mandarin hotword task. The benchmark
relies on synthetic audio and automated or model-assisted validation;
future work will extend the evaluation to human-validated recorded
conversations.

\clearpage
\bibliographystyle{IEEEbib}
\bibliography{references}

@inproceedings{mohebbi2023,
  author = {Mohebbi, H. and Chrupa{\l}a, G. and Zuidema, W. and Alishahi, A.},
  title = {Homophone Disambiguation Reveals Patterns of Context Mixing in Speech Transformers},
  booktitle = {Proc. EMNLP},
  year = {2023},
  pages = {8249--8260},
  doi = {10.18653/v1/2023.emnlp-main.513},
  url = {https://aclanthology.org/2023.emnlp-main.513/}
}

@inproceedings{dancer,
  author = {Wang, Y.-C. and Wang, H.-W. and Yan, B.-C. and Lin, C.-H. and Chen, B.},
  title = {{DANCER}: Entity Description Augmented Named Entity Corrector for Automatic Speech Recognition},
  booktitle = {Proc. LREC-COLING},
  pages = {4333--4342},
  year = {2024},
  url = {https://aclanthology.org/2024.lrec-main.387/}
}

@inproceedings{doprompts,
  author = {Yang, C.-K. and Huang, K.-P. and Lee, H.-Y.},
  title = {Do Prompts Really Prompt? Exploring the Prompt Understanding Capability of {Whisper}},
  booktitle = {Proc. IEEE SLT},
  year = {2024},
  url = {https://arxiv.org/abs/2406.05806}
}

@article{profasr,
  author = {Piskala, D. B.},
  title = {{ProfASR-Bench}: A Benchmark for Context-Conditioned {ASR} in High-Stakes Professional Speech},
  journal = {arXiv:2512.23686},
  year = {2025},
  url = {https://arxiv.org/abs/2512.23686}
}

@article{domainprompt,
  author = {Li, Y. and Wu, Y. and Li, J. and Liu, S.},
  title = {Prompting Large Language Models for Zero-Shot Domain Adaptation in Speech Recognition},
  journal = {arXiv:2306.16007},
  year = {2023},
  url = {https://arxiv.org/abs/2306.16007}
}

@inproceedings{deepcontext,
  author = {Pundak, G. and Sainath, T. N. and Prabhavalkar, R. and Kannan, A. and Zhao, D.},
  title = {Deep Context: End-to-End Contextual Speech Recognition},
  booktitle = {Proc. IEEE SLT},
  pages = {418--425},
  year = {2018},
  url = {https://research.google/pubs/deep-context-end-to-end-contextual-speech-recognition/}
}

@article{seedasr,
  author = {Bai, Y. and others},
  title = {{Seed-ASR}: Understanding Diverse Speech and Contexts with {LLM}-based Speech Recognition},
  journal = {arXiv:2407.04675},
  year = {2024},
  url = {https://arxiv.org/abs/2407.04675}
}

@inproceedings{aishell3,
  author = {Shi, Y. and Bu, H. and Xu, X. and Zhang, S. and Li, M.},
  title = {{AISHELL-3}: A Multi-Speaker Mandarin {TTS} Corpus},
  booktitle = {Proc. Interspeech},
  year = {2021},
  url = {https://www.isca-archive.org/interspeech_2021/shi21c_interspeech.html}
}

@misc{vctk,
  author = {Veaux, C. and Yamagishi, J. and MacDonald, K.},
  title = {{CSTR VCTK} Corpus: English Multi-speaker Corpus for {CSTR} Voice Cloning Toolkit (version 0.92)},
  howpublished = {Edinburgh DataShare},
  year = {2019},
  doi = {10.7488/ds/2645},
  url = {https://doi.org/10.7488/ds/2645}
}

@inproceedings{vllm,
  author = {Kwon, W. and others},
  title = {Efficient Memory Management for Large Language Model Serving with {PagedAttention}},
  booktitle = {Proc. ACM SOSP},
  year = {2023},
  url = {https://arxiv.org/abs/2309.06180}
}

@article{firered2,
  author = {Xu, K. and others},
  title = {{FireRedASR2S}: A State-of-the-Art Industrial-Grade All-in-One Automatic Speech Recognition System},
  journal = {arXiv:2603.10420},
  year = {2026},
  url = {https://arxiv.org/abs/2603.10420}
}

@article{sensevoice,
  author = {An, K. and others},
  title = {{FunAudioLLM}: Voice Understanding and Generation Foundation Models for Natural Interaction Between Humans and {LLMs}},
  journal = {arXiv:2407.04051},
  year = {2024},
  url = {https://arxiv.org/abs/2407.04051}
}

@article{vibevoiceasr,
  author = {Peng, Z. and others},
  title = {{VibeVoice-ASR} Technical Report},
  journal = {arXiv:2601.18184},
  year = {2026},
  url = {https://arxiv.org/abs/2601.18184}
}

@inproceedings{aishell1,
  author = {Bu, H. and Du, J. and Na, X. and Wu, B. and Zheng, H.},
  title = {{AISHELL-1}: An Open-Source Mandarin Speech Corpus and a Speech Recognition Baseline},
  booktitle = {Proc. Oriental COCOSDA},
  year = {2017},
  url = {https://arxiv.org/abs/1709.05522}
}

@inproceedings{librispeech,
  author = {Panayotov, V. and Chen, G. and Povey, D. and Khudanpur, S.},
  title = {{LibriSpeech}: An {ASR} Corpus Based on Public Domain Audio Books},
  booktitle = {Proc. ICASSP},
  pages = {5206--5210},
  year = {2015},
  doi = {10.1109/ICASSP.2015.7178964},
  url = {https://www.openslr.org/12}
}

@article{cosyvoice2,
  author = {Du, Z. and others},
  title = {{CosyVoice 2}: Scalable Streaming Speech Synthesis with Large Language Models},
  journal = {arXiv:2412.10117},
  year = {2024},
  url = {https://arxiv.org/abs/2412.10117}
}

@inproceedings{whisper,
  author = {Radford, A. and Kim, J. W. and Xu, T. and Brockman, G. and McLeavey, C. and Sutskever, I.},
  title = {Robust Speech Recognition via Large-Scale Weak Supervision},
  booktitle = {Proc. ICML},
  pages = {28492--28518},
  year = {2023},
  volume = {202},
  series = {PMLR},
  url = {https://proceedings.mlr.press/v202/radford23a.html}
}

@inproceedings{salm,
  author = {Chen, Z. and others},
  title = {{SALM}: Speech-Augmented Language Model with In-Context Learning for Speech Recognition and Translation},
  booktitle = {Proc. ICASSP},
  year = {2024},
  pages = {13521--13525},
  doi = {10.1109/ICASSP48485.2024.10447553},
  url = {https://arxiv.org/abs/2310.09424}
}

@inproceedings{speechllm_biasing,
  author = {Gong, X. and Lv, A. and Wang, Z. and Qian, Y.},
  title = {Contextual Biasing Speech Recognition in Speech-enhanced Large Language Model},
  booktitle = {Proc. Interspeech},
  year = {2024},
  pages = {257--261},
  doi = {10.21437/Interspeech.2024-965},
  url = {https://www.isca-archive.org/interspeech_2024/gong24b_interspeech.html}
}

@inproceedings{hori2021context,
  author = {Hori, T. and Moritz, N. and Hori, C. and {Le Roux}, J.},
  title = {Advanced Long-Context End-to-End Speech Recognition Using Context-Expanded Transformers},
  booktitle = {Proc. Interspeech},
  pages = {2097--2101},
  year = {2021},
  doi = {10.21437/Interspeech.2021-1643}
}

@article{mu2026mars,
  author = {Mu, B. and Liu, H. and Xue, H. and Wei, K. and Xie, L.},
  title = {Hearing More with Less: Multi-Modal Retrieval-and-Selection Augmented Conversational {LLM}-Based {ASR}},
  journal = {Proc. AAAI},
  volume = {40},
  number = {38},
  pages = {32519--32527},
  year = {2026},
  doi = {10.1609/aaai.v40i38.40528}
}

@article{zheng2026multilevel,
  author = {Zheng, J. and Cheng, G. and Wang, X. and Zhao, Q. and Yan, Y.},
  title = {Multilevel contextual prompting for conversational {ASR}: unifying conversation history and hotwords with speech {LLM}},
  journal = {Speech Commun.},
  volume = {183},
  pages = {103458},
  year = {2026},
  doi = {10.1016/j.specom.2026.103458}
}

@article{indiccontexteval,
  author = {Joshi, S. and others},
  title = {{IndicContextEval}: A Benchmark for Evaluating Context Utilisation in Audio Large Language Models Across 8 Indic Languages},
  journal = {arXiv:2606.19157},
  year = {2026},
  note = {Accepted at Interspeech 2026},
  doi = {10.48550/arXiv.2606.19157},
  url = {https://arxiv.org/abs/2606.19157}
}

@article{qwen3asr,
  author = {Shi, X. and others},
  title = {{Qwen3-ASR} Technical Report},
  journal = {arXiv:2601.21337},
  year = {2026},
  doi = {10.48550/arXiv.2601.21337},
  url = {https://arxiv.org/abs/2601.21337}
}

@article{contextasrbench,
  author  = {Wang, H. and others},
  title   = {{ContextASR-Bench}: A Massive Contextual Speech Recognition Benchmark},
  journal = {arXiv:2507.05727},
  year    = {2025},
  doi     = {10.48550/arXiv.2507.05727},
  url     = {https://arxiv.org/abs/2507.05727}
}

@inproceedings{conec,
  author    = {Huang, R. and others},
  title     = {{ConEC}: Earnings Call Dataset with Real-world Contexts for Benchmarking Contextual Speech Recognition},
  booktitle = {Proc. LREC-COLING},
  year      = {2024},
  pages     = {3700--3706},
  url       = {https://aclanthology.org/2024.lrec-main.328/}
}

@inproceedings{seaco,
  author = {Shi, X. and Yang, Y. and Li, Z. and Chen, Y. and Gao, Z. and Zhang, S.},
  title = {{SeACo-Paraformer}: A Non-Autoregressive {ASR} System with Flexible and Effective Hotword Customization Ability},
  booktitle = {Proc. ICASSP},
  year = {2024},
  pages = {10346--10350},
  doi = {10.1109/ICASSP48485.2024.10446106}
}

@inproceedings{slidespeech,
  author    = {Wang, H. and Yu, F. and Shi, X. and Wang, Y. and Zhang, S. and Li, M.},
  title     = {{SlideSpeech}: A Large Scale Slide-Enriched Audio-Visual Corpus},
  booktitle = {Proc. ICASSP},
  year      = {2024},
  pages     = {11076--11080},
  doi       = {10.1109/ICASSP48485.2024.10448079},
  url       = {https://doi.org/10.1109/ICASSP48485.2024.10448079}
}
\end{document}